\documentclass[runningheads]{llncs}
\usepackage[T1]{fontenc}

\usepackage{graphicx,verbatim}

\usepackage{booktabs} 
\usepackage{multirow} 
\usepackage{arydshln}
\usepackage[colorlinks=true, linkcolor=blue, citecolor=blue, urlcolor=blue]{hyperref}
\usepackage{amssymb}
\usepackage{amsmath}
\usepackage{caption}
\usepackage{marvosym}

\begin{document}
\title{MiCo: Multiple Instance Learning with Context-Aware Clustering for Whole Slide Image Analysis}
\author{Junjian Li\inst{1}\index{Li, Junjian} 
\and Jin Liu\inst{1,2}\thanks{means the corresponding author.}\index{Liu, Jin} 
\and Hulin Kuang\inst{1}\index{Kuang, Hulin} 
\and Hailin Yue\inst{1}\index{Yue, Hailin} 
\and Mengshen He\inst{1}\index{He, Mengshen} 
\and Jianxin Wang\inst{1,2}\index{Wang, Jianxin}} 
\authorrunning{J. Li et al.}
\institute{\textsuperscript{1} Hunan Provincial Key Lab on Bioinformatics, School of Computer Science and Engineering, Central South University, Changsha 410083, Hunan, China \\
\textsuperscript{2} Xinjiang Engineering Research Center of Big Data and Intelligent Software, School of Software, Xinjiang University, Urumqi 830091, China \\
 \email{liujin06@csu.edu.cn}}

\maketitle              
\begin{abstract} 
Multiple instance learning (MIL) has shown significant promise in histopathology whole slide image (WSI) analysis for cancer diagnosis and prognosis. However, the inherent spatial heterogeneity of WSIs presents critical challenges, as morphologically similar tissue types are often dispersed across distant anatomical regions. Conventional MIL methods struggle to model these scattered tissue distributions and capture cross-regional spatial interactions effectively. To address these limitations, we propose a novel Multiple instance learning framework with Context-Aware Clustering (MiCo), designed to enhance cross-regional intra-tissue correlations and strengthen inter-tissue semantic associations in WSIs. MiCo begins by clustering instances to distill discriminative morphological patterns, with cluster centroids serving as semantic anchors. To enhance cross-regional intra-tissue correlations, MiCo employs a Cluster Route module, which dynamically links instances of the same tissue type across distant regions via feature similarity. These semantic anchors act as contextual hubs, propagating semantic relationships to refine instance-level representations. To eliminate semantic fragmentation and strengthen inter-tissue semantic associations, MiCo integrates a Cluster Reducer module, which consolidates redundant anchors while enhancing information exchange between distinct semantic groups. Extensive experiments on two challenging tasks across nine large-scale public cancer datasets demonstrate the effectiveness of MiCo, showcasing its superiority over state-of-the-art methods. The code is available at https://github.com/junjianli106/MiCo.

\end{abstract}
\section{Introduction}
Whole Slide Images (WSIs) provide comprehensive and detailed representations of cellular morphology and the tumor microenvironment, both of which are crucial for accurate diagnostic evaluation and prognostic assessment \cite{cruz2014automatic,das2018multiple,zarella2019practical,lee2023recent}. However, histopathological WSIs often exhibit extremely high resolutions, reaching up to 100,000 $\times$ 100,000 pixels. The high resolution of WSIs, along with the absence of pixel-level annotations, presents significant challenges in their analysis and modeling \cite{lipkova2022deep,qu2022towards,li2022darc,li2025ca2cl}. Although Multiple Instance Learning (MIL) has emerged as a paradigm for weakly supervised WSI analysis \cite{chen2021whole,xiang2023exploring}, existing methods still struggle to handle the spatial complexity of WSIs.

This challenge manifests through two interrelated phenomena: At the macroscopic level, pathologically significant tissue structures of identical type often exhibit discontinuous distributions across distant regions. For example, lymphovascular tumor thrombi may extend from peritumoral regions to distant vascular systems \cite{lia2022diagnostic}, while perineural invasion foci can spread along nerve bundles across multiple anatomical planes \cite{medvedev2021perineural}. Microscopically, these distributed structures interact with heterogeneous cellular microenvironments, forming complex pathological networks that require simultaneous modeling of local discriminative features and cross-regional semantic dependencies.

Despite significant progress, existing MIL methods still struggle to model long-range dependencies between pathologically related regions that are morphologically distant, a challenge arising from the inherent spatial heterogeneity of WSIs. For instance, traditional attention-based MIL methods (e.g., AMIL \cite{ilse2018attention}) leverage attention mechanisms to dynamically identify and aggregate critical instances, which tend to focus on instance-level feature learning but fail to capture the semantic relationships between distant pathological regions. Clustering-based approaches, such as DeepAttnMISL \cite{yao2020whole}, learn patch dependencies within clusters but fail to capture contextual interactions among tissue regions, critical for prognosis. Transformer-based methods (e.g., TransMIL \cite{shao2021transmil}) use transformers to learn the global contextual information of WSIs but treat all interactions as homogeneous, failing to capture dynamic heterogeneity interactions across the tumor microenvironment. Graph-based methods (e.g., PatchGCN \cite{chen2021whole}) rely on fixed neighborhood definitions, while hierarchical methods like HVTSurv \cite{shao2023hvtsurv} restrict local learning to pre-determined sub-regions, both inadequate for modeling tissue-level morphological continuities across spatial distributions. 

To overcome these limitations, we propose MiCo, a novel Multiple Instance Learning framework with Context-Aware Clustering. MiCo employs a Cluster Route module to enhance intra-tissue semantic associations by aggregating and propagating information from dispersed patches of the same tissue type, refining instance-level representations. Additionally, MiCo integrates a Cluster Reducer module to eliminate semantic fragmentation and strengthen inter-tissue semantic associations, consolidating semantically redundant anchors while preserving pathological diversity. Extensive experiments on two challenging tasks across nine large-scale public cancer datasets validate MiCo's effectiveness, showcasing its superior performance over state-of-the-art methods in survival prediction and cancer subtyping tasks.

\begin{figure*}[t]
  \centering
    \includegraphics[scale=0.35]{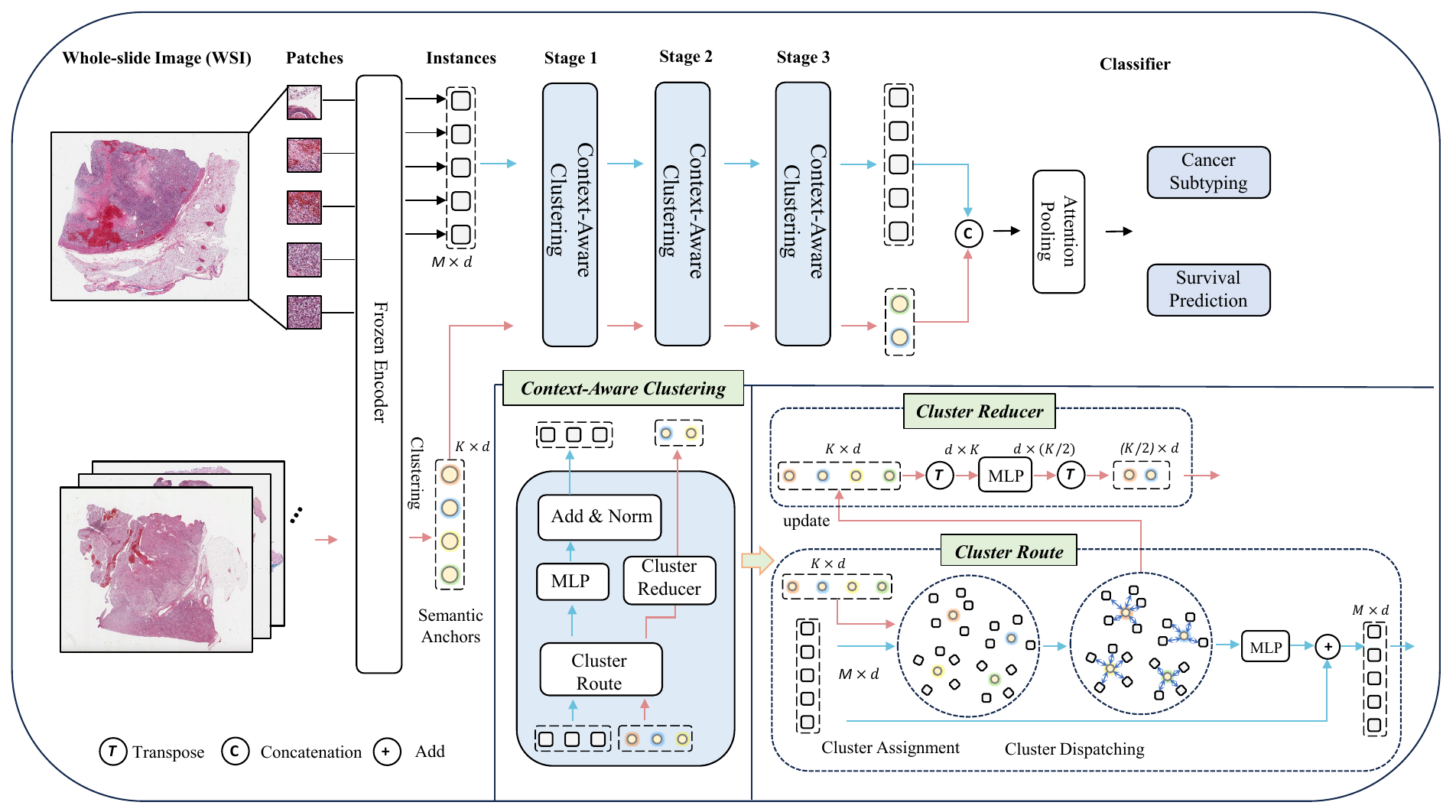}
  \caption{Overview of MiCo. MiCo consists of multi-layered context-aware clustering modules. Each module is organized by a Cluster Route module, which aggregates and propagates semantic information, and a Cluster Reducer module, which consolidates redundant anchors while enhancing information exchange between semantically distinct anchors.}
  \label{overview}
\end{figure*}

\section{Methodology}
\subsection{Overview of MiCo} 
To address the challenges of spatial heterogeneity in histopathological WSIs, we propose MiCo, a novel Multiple Instance Learning framework with Context-Aware Clustering. As illustrated in Fig. \ref{overview}, MiCo consists of multi-layered context-aware clustering modules, each organized by two core components: the Cluster Route (CluRoute) module and the Cluster Reducer (CluReducer) module. MiCo begins by clustering instances to distill discriminative morphological patterns, with cluster centroids serving as semantic anchors. The CluRoute dynamically links instances of the same tissue type across distant regions based on feature similarity, aggregating and propagating semantic information to refine instance-level representations. This process enhances cross-regional intra-tissue correlations by establishing tissue relationships that are independent of anatomical boundaries. The CluReducer consolidates redundant anchors while facilitating information exchange between distinct semantic groups, thereby eliminating semantic fragmentation and strengthening inter-tissue semantic associations. These components work synergistically to propagate contextual information across regions, refine instance-level representations through semantic aggregation, and generate robust bag-level features for downstream diagnostic and prognostic tasks.

\subsection{WSI Preprocessing}
Given a WSI \(\boldsymbol{W}_{j}\), we first crop the tissue regions into non-overlapping patches \(\left\{p_{j,m}\right\}_{m=1}^{M}\). Each patch is encoded into a feature vector \(h_{j,m} \in \mathbb{R}^{d}\) using the TITAN \cite{ding2024titan}, a multimodal whole-slide foundation model pre-trained with self-supervised and vision-language objectives. This process produces a feature matrix \(H_{j} \in \mathbb{R}^{M \times d}\) for each WSI. To simplify notation, we drop the subscript \(j\). To initialize semantic anchors, we apply K-means clustering on the training instances, obtaining cluster centers \(S = \left\{s_{k}\right\}_{k=1}^{K}\). These centers serve as semantic anchors representing distinctive tissue patterns.

\subsection{Context-Aware Clustering} 
Morphologically similar tissue types are often dispersed across distant anatomical regions, making it challenging to capture their contextual relationships and semantic consistency. MiCo addresses this challenge through its multi-layered context-aware clustering modules, which consist of two core components: the Cluster Route and the Cluster Reducer. The Cluster Route aggregates and propagates semantic information by linking instances of the same tissue type across distant regions, while the Cluster Reducer consolidates redundant anchors and facilitates information exchange between distinct semantic groups.

\noindent\textbf{Cluster Route.} 
Given a WSI containing \( M \) instances, each patch \( p_m \) is represented by a feature embedding \( h_m \in \mathbb{R}^d \). To establish cross-regional intra-tissue correlations, we compute the cosine similarity between each instance feature \( h_i \) and \( K \) learnable semantic anchors \( \{s_k\}_{k=1}^K \): 
\begin{equation}
A_{l(m,k)} = \frac{h_m^\top s_k}{|h_m| \cdot |s_k|},
\label{eq1}
\end{equation}
where \( A_l \in \mathbb{R}^{M \times K} \) quantifies the semantic alignment between instances and semantic anchors.

The argmax operation ensures that each instance is assigned to its most relevant semantic anchor, preserving semantic consistency across pathological structures. However, argmax is non-differentiable, which disrupts gradient flow to the anchors. To address this, we use a straight-through estimator \cite{van2017neural} to enable instance-to-anchor assignments while maintaining gradient propagation:
\begin{equation}
\hat{A}_l = \text{one-hot}\left(\arg\max(A_l)\right) + A_l - \text{sg}(A_l),
\label{eq2}
\end{equation}
where \( \text{sg}(\cdot) \) denotes the stop-gradient operator. During forward propagation, \( \hat{A}_l \in \mathbb{R}^{M \times K} \) is a binary matrix where each row is one-hot encoded, assigning each instance to its most similar cluster. During backward propagation, gradients bypass the non-differentiable \( \arg\max \) operation and flow through \( A_l \), enabling end-to-end training.  

For each semantic anchor \( s_{k} \), we aggregate the features of its assigned instances to generate a context-aware representation:
\begin{equation}
\begin{aligned}
\tilde{s}_k = \frac{1}{N_k} \sum_{m=1}^M \hat{A}_{l(m,k)} \cdot h_m, \quad N_k = \sum_{m=1}^M \hat{A}_{l(m,k)}. 
\end{aligned}
\label{eq3}
\end{equation}
where \( \tilde{s}_k \in \mathbb{R}^d \) captures shared morphological semantics (e.g., tumor stroma or immune infiltration) within the anchor group, and $N_{k}$ denotes the number of instances assigned to the $k$-th anchor. This aggregation step unifies spatially dispersed instances sharing similar semantics, directly mitigating spatial heterogeneity. We collect all \( \tilde{s}_k \) into a matrix \( \tilde{S} \in \mathbb{R}^{K \times d} \), which serves as the input to the Cluster Reducer module. 

To propagate inter-region context, we align each instance feature \( h_m \) with its assigned semantic anchor semantics:  
\begin{equation}
\begin{aligned}
h_m^{'} = h_m + MLP(h_m +  \hat{A}_{l(m,k)} \cdot \tilde{s}_k).
\end{aligned}
\label{eq4}
\end{equation}
The updated feature \( h_m^{'} \) combines instance-level features with cross-regional semantic context, creating a unified representation of spatially dispersed tissue structures. This enriched feature is then passed to the next context-aware clustering module for further refinement, enabling the model to progressively capture complex anatomical relationships.

\noindent\textbf{Cluster Reducer.} 
To reduce semantic redundancy, the Cluster Reducer merges anchors with similar characteristics by modeling inter-anchor relationships. Given initial semantic anchors \( \tilde{S} \in \mathbb{R}^{K \times d} \), we transpose \( \tilde{S} \) to \( \tilde{S}^\top \in \mathbb{R}^{d \times K} \) and apply a MLP to model nonlinear interactions:  
\begin{equation}
\tilde{S}^{'} = MLP(\tilde{S}^\top),
\label{eq5}
\end{equation}
where \( \tilde{S}^{'} \in \mathbb{R}^{d \times K/2} \) represents the refined anchors after redundancy reduction. The MLP captures semantic dependencies, such as merging scattered tumor-infiltrating lymphocyte clusters, by consolidating semantically similar anchors. Transposing \( \tilde{S}^{'} \) back to \( \tilde{S} \in \mathbb{R}^{K/2 \times d} \) yields compact and meaningful anchors that preserve essential tissue characteristics.

\section{Experiments and Results} 
\subsection{Datasets} 
\noindent\textbf{Survival Prediction.} We use seven publicly available cancer datasets from TCGA\footnote{https://portal.gdc.cancer.gov} (\textbf{BLCA}, \textbf{BRCA}, \textbf{GBMLGG}, \textbf{HNSC}, \textbf{KIRC}, \textbf{KIRP}, and \textbf{LUAD}) in our experiments. These datasets collectively include data from 3,523 patients and 4,091 H\&E diagnostic WSIs.

\noindent\textbf{Cancer Subtyping.} We conduct comparative experiments on two challenging public datasets: \textbf{TCGA-BRCA} and \textbf{TCGA-NSCLC}. The TCGA-BRCA dataset contains 1,034 H\&E slides of two invasive cancer subtypes: invasive ductal carcinoma (IDC) and invasive lobular carcinoma (ILC). The TCGA-NSCLC dataset includes 1,030 H\&E slides from two subtypes: Lung Squamous Cell Carcinoma (TCGA-LUSC) and Lung Adenocarcinoma (TCGA-LUAD).

\subsection{Implementation Details} 
We compare MiCo with eight state-of-the-art methods: AMIL \cite{ilse2018attention}, TransMIL \cite{shao2021transmil}, DeepGraphConv \cite{li2018graph}, Patch-GCN \cite{chen2021whole}, ILRA \cite{xiang2023exploring}, HVTSurv \cite{shao2023hvtsurv}, RRTMIL \cite{tang2024feature}, and WiKG \cite{li2024dynamic}. WSIs are preprocessed using TITAN \cite{ding2024titan} to extract 448 $\times$ 448 patch features at 20 $\times$ magnification. All experiments adopt 4-fold cross-validation, with datasets split into training, validation, and test sets at a ratio of 60:15:25. The semantic anchor number is set to 64. Training configurations are unified across methods: 200 epochs, batch size 1, learning rate 2e-4 (Ranger optimizer \cite{wright2019ranger}), early stop 8, and gradient accumulation every 2 steps. Survival prediction performance is evaluated using the Concordance Index (C-Index) with standard deviation (std), while cancer subtyping is assessed via Accuracy (ACC), F1-score (F1), and Area Under the Curve (AUC) metrics. 

\begin{table}[t!]
    \centering
    \caption{Survival prediction performance comparison across seven cancer types.}
    \label{results_survival}
     \resizebox{1.0\textwidth}{!}{
        \begin{tabular}{c|ccccccc|c}
        \hline
        Models & BLCA & BRCA & GBMLGG & HNSC  & KIRC & KIRP & LUAD & MEAN \\
        \hline
        AMIL \cite{ilse2018attention} &  0.551$_{0.10}$ & 0.576$_{0.11}$ & 0.736$_{0.15}$ & 0.590$_{0.02}$ & 0.681$_{0.05}$ & 0.778$_{0.04}$ & 0.584$_{0.04}$ & 0.642 \\
        TransMIL \cite{shao2021transmil} & 0.610$_{0.07}$ & 0.555$_{0.04}$ & 0.760$_{0.14}$ & 0.565$_{0.02}$ & 0.694$_{0.05}$ & 0.773$_{0.07}$ & 0.585$_{0.09}$ & 0.649 \\
        DeepGraphConv \cite{li2018graph} & 0.563$_{0.08}$ & 0.581$_{0.10}$ & 0.739$_{0.14}$ & 0.591$_{0.02}$ & 0.681$_{0.02}$ & 0.738$_{0.04}$ & 0.581$_{0.02}$ & 0.639 \\
        PatchGCN  \cite{chen2021whole} & 0.607$_{0.05}$ & 0.600$_{0.12}$ & 0.736$_{0.16}$ & 0.584$_{0.01}$ & 0.694$_{0.05}$ & 0.783$_{0.05}$ & 0.582$_{0.04}$ & 0.655 \\
        ILRA  \cite{xiang2023exploring} & 0.612$_{0.04}$ & 0.551$_{0.07}$ & 0.722$_{0.16}$ & 0.600$_{0.01}$ & 0.663$_{0.04}$ & 0.688$_{0.05}$ & 0.590$_{0.04}$ & 0.632 \\
        HVTSurv \cite{shao2023hvtsurv} & 0.606$_{0.05}$ & 0.579$_{0.09}$ & 0.792$_{0.01}$ & 0.575$_{0.03}$ & 0.692$_{0.04}$ & 0.778$_{0.06}$ & 0.591$_{0.03}$ & 0.658 \\
        RRTMIL \cite{tang2024feature} & 0.588$_{0.05}$ & 0.542$_{0.08}$ & 0.738$_{0.16}$ & 0.597$_{0.02}$ & 0.699$_{0.04}$ & 0.770$_{0.06}$ & 0.579$_{0.04}$ & 0.644 \\
        WiKG  \cite{li2024dynamic} & 0.607$_{0.07}$ & 0.582$_{0.09}$ & 0.752$_{0.16}$ & 0.579$_{0.02}$ & 0.704$_{0.06}$ & 0.763$_{0.09}$ & 0.595$_{0.06}$ & 0.654 \\
        \textbf{MiCo} & \textbf{0.619$_{0.03}$} & \textbf{0.608$_{0.12}$} & \textbf{0.813$_{0.02}$} & \textbf{0.605$_{0.01}$} & \textbf{0.715$_{0.05}$} & \textbf{0.800$_{0.07}$} & \textbf{0.604$_{0.04}$} & \textbf{0.680} \\
        \hline
        \end{tabular}
    }
\end{table}

\begin{table}[t!]
    \centering
    \caption{Cancer subtyping performance comparison across two cancer types.}
    \label{results_subtype}
    \resizebox{1.0\textwidth}{!}{
        \begin{tabular}{cccccccccccc}
            \toprule
            \multirow{2}{*}{Method} & \multicolumn{3}{c}{TCGA-BRCA} & \phantom{a} & \multicolumn{3}{c}{TCGA-NLCSC} & \phantom{a} & \multicolumn{3}{c}{MEAN} \\
            \cmidrule{2-4} \cmidrule{6-8} \cmidrule{10-12} 
                                    & ACC  & F1  & AUC && ACC & F1 & AUC  && ACC & F1 & AUC  \\ 
            \midrule
            AMIL  \cite{ilse2018attention}        & 0.909$_{0.01}$  & 0.858$_{0.02}$ & 0.948$_{0.01}$  && 0.923$_{0.02}$  & 0.923$_{0.02}$ & 0.979$_{0.01}$ && 0.916 & 0.890 & 0.964 \\
            TransMIL \cite{shao2021transmil}  & 0.906$_{0.01}$  & 0.857$_{0.02}$ & 0.945$_{0.02}$  && 0.919$_{0.02}$  & 0.919$_{0.02}$ & 0.977$_{0.02}$ && 0.913 & 0.889 & 0.961 \\
            DeepGraphConv \cite{li2018graph} & 0.913$_{0.01}$  & 0.859$_{0.01}$ & 0.947$_{0.01}$  && 0.913$_{0.02}$  & 0.913$_{0.02}$ & 0.969$_{0.02}$ && 0.913 & 0.886 & 0.958 \\
            PatchGCN \cite{chen2021whole} &   0.917$_{0.01}$  & 0.871$_{0.02}$ & 0.944$_{0.02}$  && 0.919$_{0.02}$  & 0.919$_{0.02}$ & 0.979$_{0.01}$ && 0.918 & 0.895 & 0.962 \\
            ILRA   \cite{xiang2023exploring}       & 0.899$_{0.01}$  & 0.836$_{0.03}$ & 0.931$_{0.02}$  && 0.918$_{0.02}$  & 0.918$_{0.02}$ & 0.973$_{0.02}$ && 0.909 & 0.885 & 0.952 \\
            RRTMIL \cite{tang2024feature}     & 0.910$_{0.01}$  & 0.855$_{0.01}$ & 0.945$_{0.02}$  && 0.915$_{0.02}$  & 0.915$_{0.02}$ & 0.976$_{0.01}$ && 0.913 & 0.890 & 0.961 \\
            WiKG  \cite{li2024dynamic}        & 0.915$_{0.02}$  & 0.859$_{0.03}$ & 0.948$_{0.01}$  && 0.920$_{0.03}$  & 0.920$_{0.03}$ & 0.980$_{0.01}$ && 0.917 & 0.889 & 0.964 \\
            \textbf{MiCo}             & \textbf{0.922$_{0.01}$}  & \textbf{0.875$_{0.02}$} & \textbf{0.952$_{0.01}$}  && \textbf{0.931$_{0.02}$}  & \textbf{0.931$_{0.02}$} & \textbf{0.981$_{0.01}$} && \textbf{0.927} & \textbf{0.903} & \textbf{0.967} \\
            \bottomrule
        \end{tabular}
    }
\end{table}

\subsection{Results and Discussion}
\noindent\textbf{Survival Prediction.}
The experimental results demonstrate the superior performance of MiCo across seven cancer types for survival prediction tasks. As shown in Table \ref{results_survival}, MiCo achieves a mean C-index of 0.680, outperforming all other methods. This performance underscores the effectiveness of MiCo's context-aware clustering in addressing the challenges posed by spatial heterogeneity in histopathological WSIs. MiCo's ability to model cross-regional dependencies and enhance semantic associations across dispersed tissue regions enables it to capture complex morphological patterns and improve survival prediction accuracy.

\begin{figure}[t!]
\centering
\includegraphics[scale=0.042]{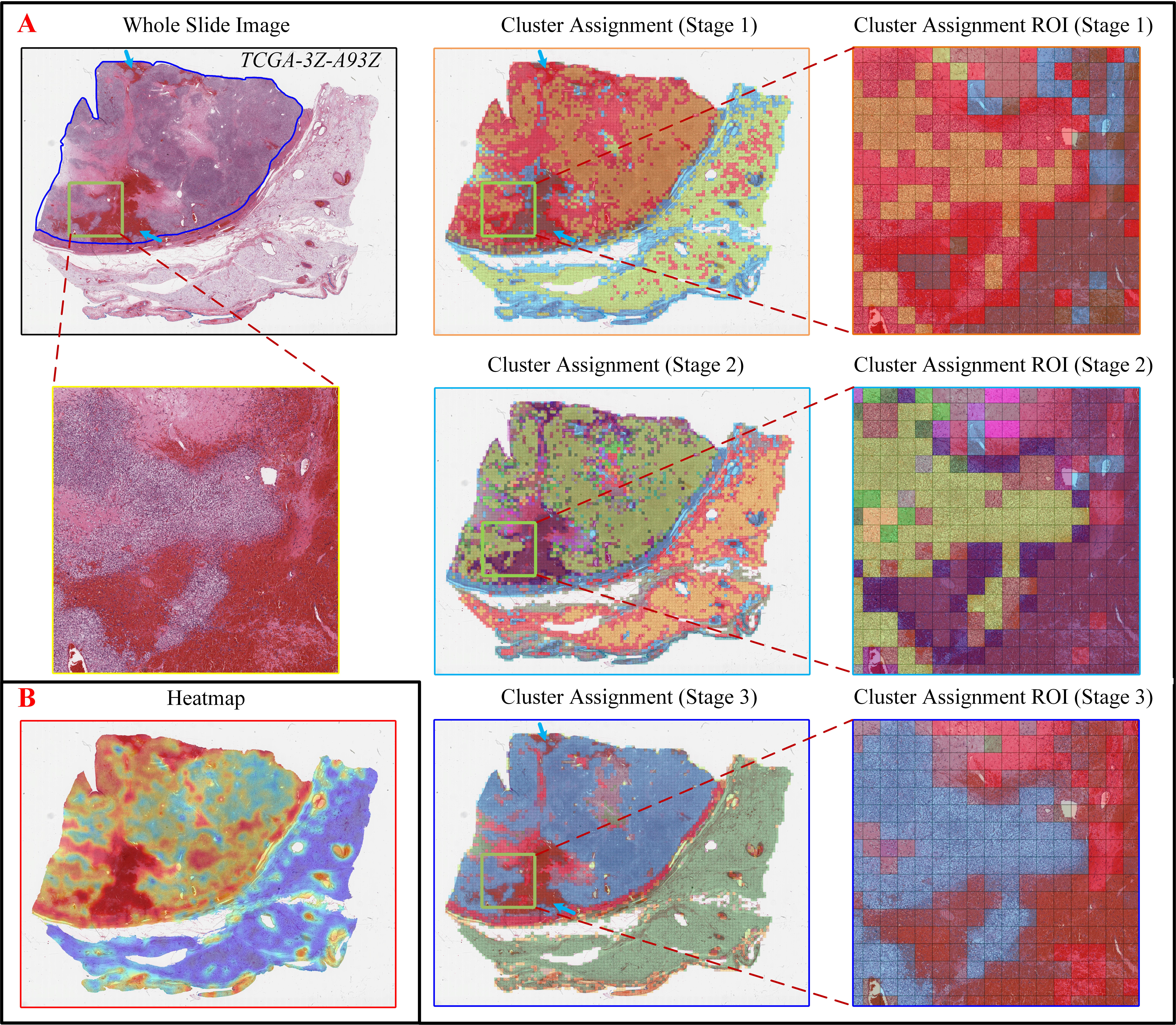} 
\caption{ 
\textbf{A.} Qualitative results of cluster assignment, showcasing WSI thumbnails, cluster assignment results, and corresponding regions of interest (ROIs) at different stages. Blue outlines denote ground truth tumor regions, while areas assigned to the same cluster are displayed in identical colors. \textbf{B.} Interpretability analysis of MiCo, with red regions in the heatmaps indicating areas of high attention.}
\label{heatmap}
\end{figure}

\noindent\textbf{Cancer Subtyping.}
The experimental results for cancer subtyping, as presented in Table \ref{results_subtype}, demonstrate MiCo's superior performance across two main datasets, TCGA-BRCA and TCGA-NLCSC. MiCo achieves state-of-the-art results in terms of accuracy, F1 score, and AUC, outperforming all other methods. With a mean ACC of 0.927, MiCo significantly surpasses the best-performing baseline, PatchGCN, which achieves a mean ACC of 0.918. These results highlight MiCo's ability to accurately classify cancer subtypes by leveraging its context-aware clustering mechanism and dynamic semantic anchors, effectively alleviating the challenges of spatial heterogeneity and semantic fragmentation in histopathological WSIs.

\begin{figure}[t!]
  \centering
  \begin{minipage}[c]{0.44\linewidth} 
    \centering
    \includegraphics[width=\linewidth]{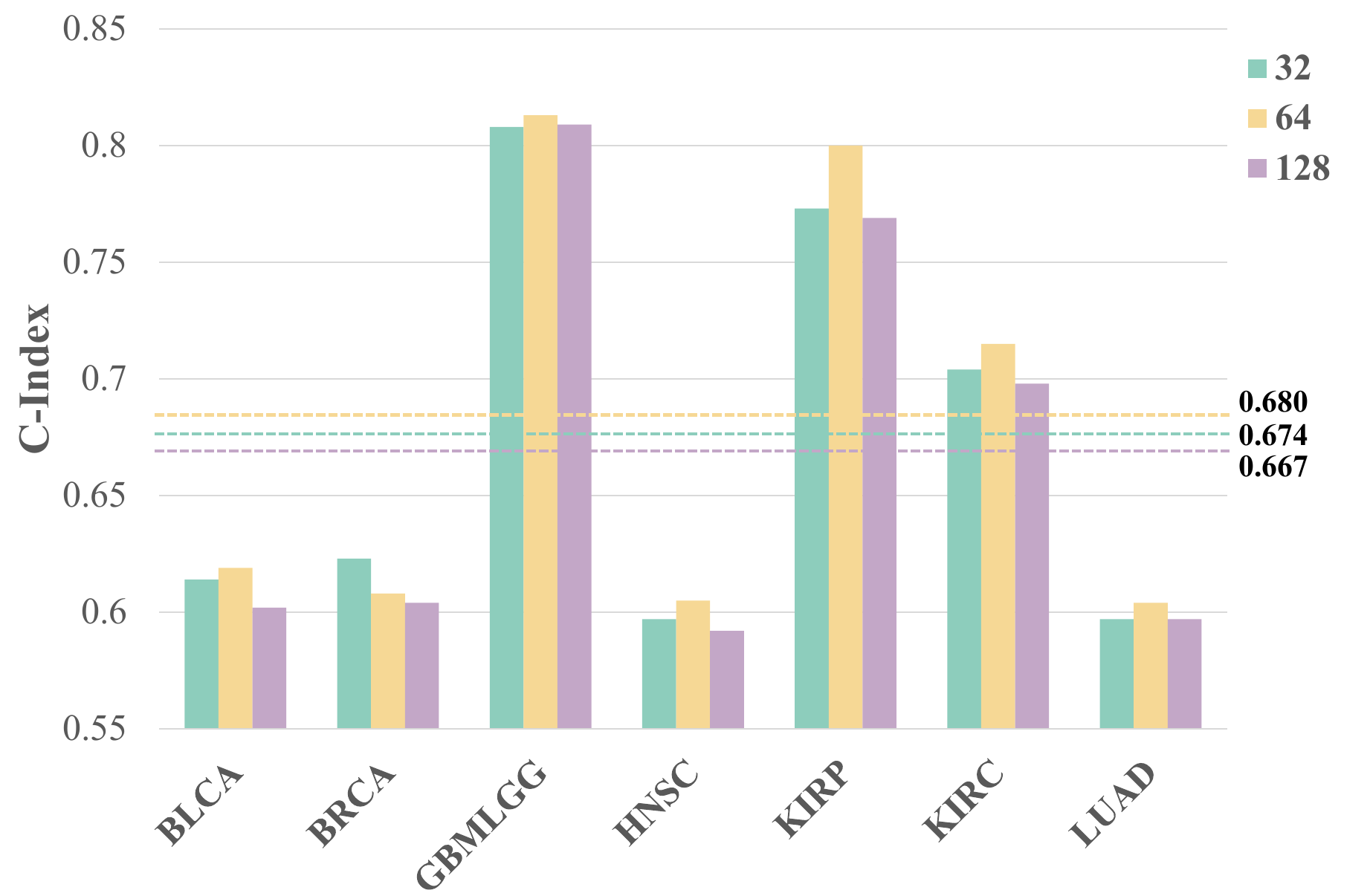}
    \caption{
      Analysis of semantic anchors' impact on survival prediction.
    }
    \label{ClusterNum}
  \end{minipage}
  \hfill 
  \begin{minipage}[c]{0.5\linewidth} 
    \centering
    \captionof{table}{Average results of the ablation study for MiCo.} 
    \label{results_ablation}
    \resizebox{\linewidth}{!}{ 
      \begin{tabular}{ccccc}
        \toprule
        \multirow{2}{*}{Method} & Survival Prediction & \multicolumn{3}{c}{Cancer Subtyping} \\
        \cmidrule{2-2} \cmidrule{3-5}
                                & C-Index  & ACC  & F1  & AUC  \\ 
        \midrule
        w/o Semantic Anchors Init              & 0.668  & 0.922 & 0.896 & 0.964 \\
        w/o CluReducer          & 0.659  & 0.921 & 0.897 & 0.965 \\
        w/o CluRoute           & 0.656  & 0.910 & 0.890 & 0.955 \\
        \textbf{MiCo}           & \textbf{0.680}  & \textbf{0.927} & \textbf{0.903} & \textbf{0.967} \\
        \bottomrule
      \end{tabular}
    }
  \end{minipage}
\end{figure}

\noindent\textbf{Ablation Study.} To further determine the efficacy of each crucial component in MiCo, we conduct a series of ablation studies on survival prediction and cancer subtyping tasks. The average results, as summarized in Table \ref{results_ablation}, underscore the critical importance of each module in enhancing the overall performance of MiCo. The ablation experiments evaluate the impact of removing key components, including semantic anchor initialization, the CluReducer, and the CluRoute. Removing semantic anchor initialization reduces the C-index to 0.668, underscoring its importance in capturing discriminative morphological patterns. Omitting the CluReducer further decreases the C-index to 0.659, highlighting its role in consolidating redundant anchors and enhancing inter-anchor information exchange. Similarly, removing the CluRoute results in a C-index of 0.656, demonstrating its effectiveness in establishing cross-regional semantic relationships and refining instance-level representations. These results validate the critical contribution of each module to MiCo's overall performance.

\noindent\textbf{Interpretability Analysis.} As shown in Fig.\ref{heatmap}, we analyze MiCo's interpretability through cluster assignments and heatmaps. In Fig.\ref{heatmap}.A, semantically related regions are progressively merged across stages, demonstrating MiCo's ability to refine representations and enhance semantic coherence. The heatmap results further reveal strong alignment with ground truth tumor annotations in Fig.\ref{heatmap}.B, confirming MiCo's high accuracy in tumor localization. 

\noindent\textbf{Semantic Anchors Analysis.} 
As shown in Fig.\ref{ClusterNum}, we evaluate MiCo's survival prediction performance with different numbers of semantic anchors (32, 64, and 128). The mean C-index values are 0.674 (32 anchors), 0.680 (64 anchors), and 0.667 (128 anchors). The 64-anchor configuration achieves the highest mean C-index, balancing morphological diversity and redundancy. 

\section{Conclusion}
In this paper, we propose MiCo, a novel Multiple Instance Learning framework with Context-Aware Clustering, designed to tackle spatial heterogeneity in histopathological WSIs. MiCo incorporates a Cluster Route module to strengthen cross-regional intra-tissue correlations and a Cluster Reducer module to consolidate redundant anchors while promoting information exchange between distinct semantic groups. Extensive experiments on nine cancer datasets validate MiCo's superiority over state-of-the-art methods in both survival prediction and cancer subtyping tasks.

\noindent\textbf{Acknowledgement.}\footnote{The authors have no competing interests to declare that are relevant to the content of this article.} This work was supported in part by Xinjiang Uygur Autonomous Region Key R\&D program (No. 2024B03039-3), the National Natural Science Foundation of China (No. U24A20256), the Scientific Research Fund of Hunan Provincial Education Department (No. 23A0020), the Science and Technology Major Project of Changsha (No. kh2502004), the central government guides local funds for scientific and technological development of China (No. ZYYD2025QY25), the Central South University Innovation-Driven Research Programme (No. 2023CXQD018), and the Fundamental Research Funds for the Central Universities of Central South University (No. 2024ZZTS0108). This work was carried out in part using computing resources at the High-Performance Computing Center of Central South University.

\bibliographystyle{splncs04}
\bibliography{ref}
\end{document}